\documentclass[10pt,twocolumn,letterpaper]{article}

\usepackage{cvpr}              % To produce the CAMERA-READY version
\newcommand{\tb}[1]{\textcolor[rgb]{0,0,0.0}{#1}}    %tilo
\newcommand{\dat}[1]{\textcolor[rgb]{0,0,1}{#1}}
\usepackage{comment}
\usepackage{multirow}
\usepackage{array}
\usepackage{booktabs}
\usepackage{times}     % For Times font
\usepackage{multirow}
\usepackage{scrextend}
\definecolor{cvprblue}{rgb}{0.21,0.49,0.74}
\usepackage[pagebackref,breaklinks,colorlinks,allcolors=cvprblue]{hyperref}

\def\paperID{42} % *** Enter the Paper ID here
\def\confName{CVPR}
\def\confYear{2025}

\title{\vspace{-25pt}\tb{WildLive: Near Real-time Visual Wildlife Tracking onboard UAVs}\vspace{-8pt}}

\author{
Nguyen Ngoc Dat\textsuperscript{1}, 
Tom Richardson\textsuperscript{1}, 
Matthew Watson\textsuperscript{1}, 
Kilian Meier\textsuperscript{1}, \\
Jenna Kline\textsuperscript{2},
Sid Reid\textsuperscript{1}, 
Guy Maalouf\textsuperscript{3}, 
Duncan Hine\textsuperscript{1}, 
Majid Mirmehdi\textsuperscript{1}, 
Tilo Burghardt\textsuperscript{1} \\
\textsuperscript{1} {\tt\footnotesize{University of Bristol}, \footnotesize{UK}} \quad 
\textsuperscript{2} {\tt\footnotesize{Ohio State University}, \footnotesize{USA}} \quad 
\textsuperscript{3} {\tt\footnotesize{University of Southern Denmark}, \footnotesize{Denmark}}}

\begin{document}
\maketitle

%%%%%%%%% ABSTRACT
\begin{abstract}
{Live tracking of wildlife via high-resolution video processing directly onboard drones is widely unexplored and most existing solutions rely on streaming video to ground stations to support navigation. Yet, both autonomous animal-reactive flight control beyond visual line of sight and/or mission-specific individual and behaviour recognition tasks rely to some degree on this capability. In response, we introduce WildLive -- a near real-time animal detection and tracking framework for high-resolution imagery running directly onboard uncrewed aerial vehicles (UAVs). The system performs multi-animal detection and tracking at 17.81~fps for HD and 7.53~fps on 4K video streams suitable for operation during higher altitude flights to minimise animal disturbance. Our system is optimised for Jetson Orin AGX onboard hardware. It integrates the efficiency of sparse optical flow tracking and mission-specific sampling with device-optimised and proven YOLO-driven object detection and segmentation techniques. Essentially, computational resource is focused onto spatio-temporal regions of high uncertainty to significantly improve UAV processing speeds. Alongside, we introduce our WildLive dataset, which comprises 200K+ annotated animal instances across 19K+ frames from 4K UAV videos collected at the Ol Pejeta Conservancy in Kenya. All frames contain ground truth bounding boxes, segmentation masks, as well as individual tracklets and tracking point trajectories. We compare our system against current object tracking approaches including OC-SORT, ByteTrack, and SORT. Our multi-animal tracking experiments with onboard hardware confirm that near real-time high-resolution wildlife tracking is possible on UAVs whilst maintaining high accuracy levels as needed for future navigational and mission-specific animal-centric operational autonomy. We publish all source code, weights, dataset, and labels for easy utilisation by the community.}
\end{abstract}

\begin{figure}[h]
    \centering\vspace{-5pt}\hspace{-15pt}
    \includegraphics[width=240pt, height=300pt]{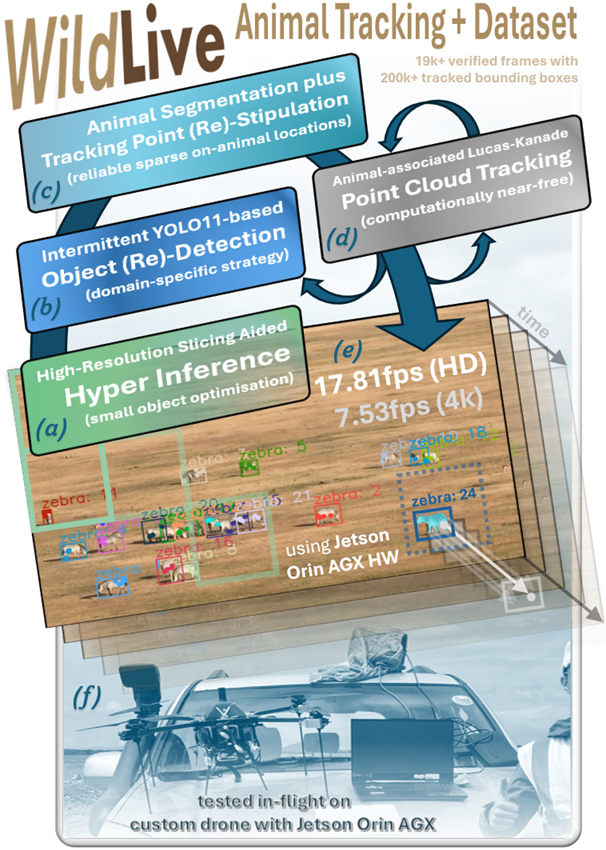}\vspace{-7pt}
    \caption{\tb{\textbf{WildLive System Overview.} Our pioneering approach integrates the efficiency of \textit{\textbf{(a)}} Slicing-aided Hyper Inference with proven YOLO-driven \textit{\textbf{(b)}} object re-detection and \textit{\textbf{(c)}} segmentation techniques. The framework exploits animal-associated \textit{\textbf{(d)}}~inexpensive Lucas-Kanade point tracks to interpolate intermittent re-detection allowing  \textit{\textbf{(e)}} high-speed HD/4K tracking directly onboard UAVs utilising \textit{\textbf{(f)}} custom drones with Jetson Orin AGX hardware.}}\vspace{-9pt}
    \label{fig:overview}
\end{figure}
    
\vspace{-11pt}
\section{\tb{Introduction and Motivation}}\vspace{-3pt}
\label{sec:intro}

{\textbf{Live Tracking of Animals via Drones.} Multi-object tracking~(MOT) \cite{motsurvey,motsurvey2} in high-resolution video streams processed live onboard drones~\cite{visualobj1} poses significant challenges when applied to wildlife monitoring~\cite{perspectives,vayssade2019automatic,wildlife_monitor,wildwing} due to environmental demands, small animal resolution~\cite{xue2022small}, platform motion, computational constraints, energy limitations and more. Yet, this capability plays a key role in enabling future animal-reactive and/or Beyond Visual Line of Sight (BVLOS) flight control for mission-specific individual~\cite{andrew2019aerial,wildbridge} and behaviour recognition~\cite{burghardt2006real,brookes2024panaf20k,deep_dive_kabr,yolo_behaviour} -- without options to involve ground control due to latency or connectivity constraints. Although first attempts~\cite{naturepaper} to build such systems exist, operation is currently limited to low resolutions without full benchmarking where datasets/code are so far not fully public\footnote{\tb{Parts of the source code for~\cite{naturepaper} is available at~\url{https://github.com/hardboy12/YOLOv7-DeepSORT}}.}}.
{While existing datasets \cite{mmla,bucktales,deep_dive_kabr} have addressed multi-object tracking from aerial perspectives or in wildlife-focused contexts, they remain limited in scope or applicability. In contrast, our dataset combines high-resolution 4K wildlife video, a drone perspective and multi-object tracking supporting both the standard MOT task as well as point tracking~\cite{tapvid,tapir}. It incorporates manually verified animal segmentation masks and individual point tracks in the ground truth. This enables robust MOT benchmarking while also opening up opportunities for Track-Any-Point~(TAP) research in challenging wildlife scenarios, making our dataset a valuable resource for both traditional and emerging tracking methodologies.}

In response, this paper proposes, benchmarks and shares with the community the near real-time MOT WildLive system~(see~Fig.~\ref{fig:overview}) suitable for advanced tracking of animals in high-resolution video streams during flight directly onboard UAVs with a Jetson Orin AGX.\vspace{5pt}

\noindent{{Our main contributions in this work are:}
\begin{enumerate} % Use Roman numerals with uppercase letters
  \item  We introduce and make publicly available\footnote{\label{note2}\tb{WildLive materials, source code and links can be found at \url{https://dat-nguyenvn.github.io/WildLive/}}} our MOT WildLive system~(see Sec.~\ref{sec:method}) for near real-time wildlife tracking optimised for and deployable directly on drones that carry an embedded Jetson Orin AGX computer. 
  \item We benchmark WildLive against suitable SOTA systems on a domain-specific tracking dataset~(see Sec.~\ref{sec:dataset}) which we introduce alongside our system. Amongst other rich annotations, it contains  200K+~tracked bounding boxes from representative, UAV-acquired video sequences recorded on site in Kenya under strict ethical oversight (see Ethics Statement). 
  \item We publish\footref{note2} our WildLive Benchmark Dataset and its ground truth information in full for reproducibility and domain-relevant comparability in this evolving field. 
%215800 bounding boxes , 908 id,19139 frames, 2 videos
 %See figure 
  %\item A Fine tune model for drone (optional)
\end{enumerate}}

%\begin{figure*}[t] % 't' positions the figure at the top of the page
%\centering
%\includegraphics[width=\textwidth]{images/technique.png} % Replace with your image file
%\caption{Novel technique CoTracker on an embedded computer }
%\label{fig:fullwidth}
%\end{figure*}

\vspace{-3pt}\section{\textbf{Paper Concept and Related Work}}\vspace{-3pt}
\label{sec:related}

{\textbf{Detection vs. Tracking.} Whilst localising the presence of objects in video, i.e. \textit{detection}, requires matching complex pixel patterns over potentially large object regions, following content across frames, i.e. \textit{tracking}, may either utilise these entire objects~\cite{bytetrack,visualobjD,sort,motsurvey} or alternatively follow discrete, potentially sparse locations on the objects~\cite{tapvid,tapir,cotracker} only. ByteTrack \cite{bytetrack}, OCSORT \cite{ocsort}, and SORT \cite{sort} are examples of recent \textit{full object trackers} computationally suitable for edge device deployment. On the other hand, deep trackers such as CoTracker~\cite{cotracker} and traditional sparse optical flow trackers such as Lucas-Kanade~(LK)~\cite{lkp} implement \textit{point tracking} capabilities. The latter are vastly cheaper computationally, but have shortcomings regarding occlusions, aperture limitations and viewpoint changes. Given that object (re)detection may not be required every frame, this offers performance headroom to combine and balance fast LK tracking with intermittent deep detection to perform light-weight, UAV-suitable wildlife tracking in high-resolution video near real-time.}

{\textbf{\textbf{Slicing Aided Hyper Inference (SAHI) and YOLO.}} To date, the YOLO~\cite{yolov5,yolo6,yolo7,ultralytics,yolo11} detector series have remained amongst the fastest deep object detectors of their time throughout their version history. To process high-resolution 4K images with YOLO without performance-crushing information loss due to downsampling, we apply the SAHI technique~\cite{sahi} in conjunction with YOLOv8~\cite{ultralytics} or YOLO11~\cite{yolo11} for optimised processing speeds beyond simple window processing without loss of accuracy. Note that all performance metrics are evaluated under this same regime including tracker benchmarking for a fair comparison and maximal utilisation of video content -- particularly given small animal sizes in most UAV-acquired footage.}

{{\textbf{\textbf{Evaluation Metrics.}}} We adopt MOTA (Multiple Object Tracking Accuracy) \cite{evalmethod1,eval2} and IDF1 (Identification F1 Score) \cite{evalmethod1} for performance measurements. These are widely used measures in the domain and allow for broad comparability with other systems. Additionally, we will be benchmarking our system with TETA~\cite{teta} in future work.}

\begin{figure}[t]
  \centering
    \centering\vspace{-2pt}
    \includegraphics[width=240pt, height=170pt]{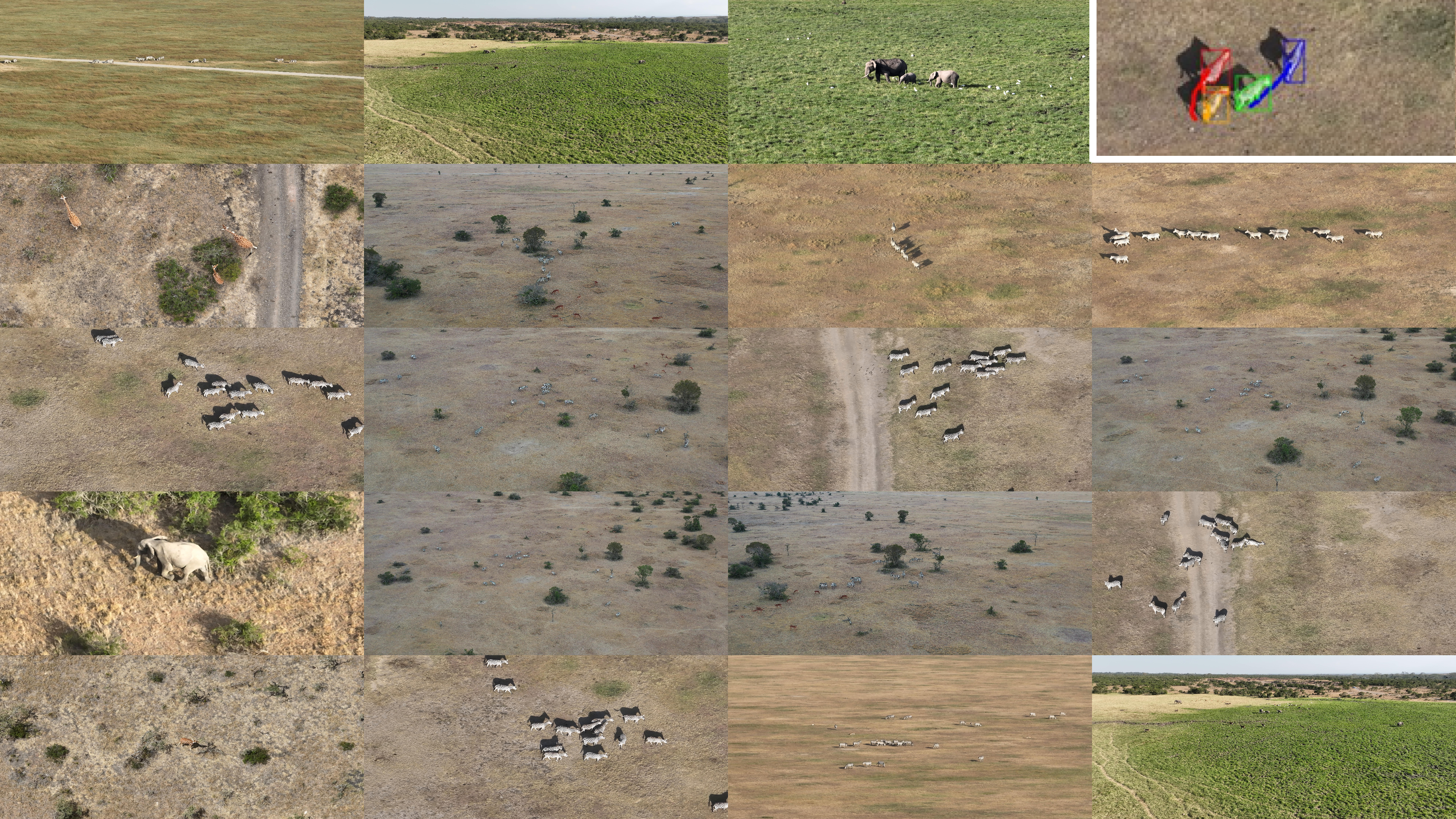}\vspace{-7pt}
   \caption{{\textbf{WildLive Benchmark Dataset Overview.} 19 representative 4K frames (each sampled from a different video) showcasing the dataset's diversity regarding altitudes, environments, species, approach angles as well as view points. The top right image shows a zoomed-in example patch with ground truth annotations of animal bounding boxes, segmentations, and tracked point trajectories.}}\vspace{-8pt}
   \label{fig:dataset}
\end{figure}

\vspace{-3pt}\section{Dataset}
\begin{figure}[t]
    \centering
    \includegraphics[width=\linewidth]{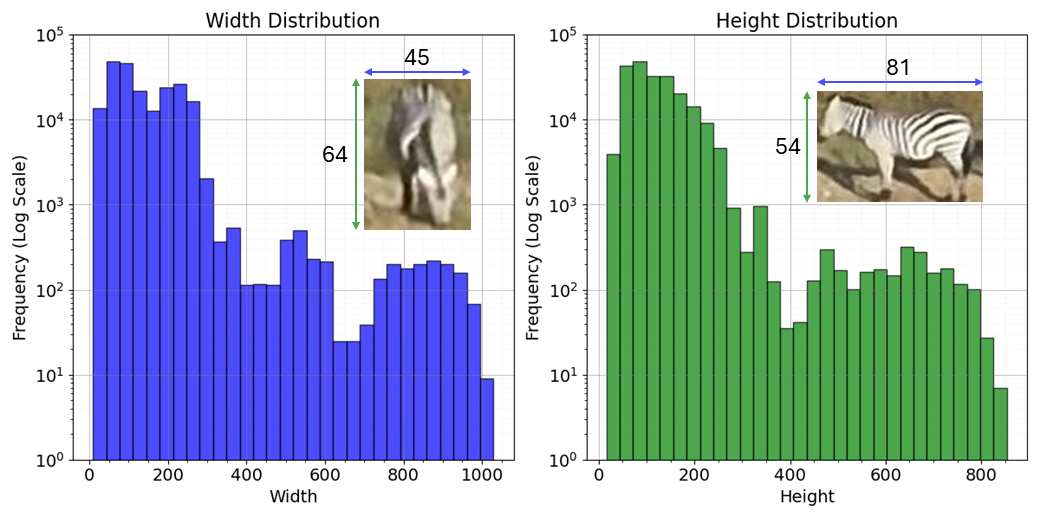}\vspace{-10pt}
    \centering
    \caption{{\textbf{Distribution of Animal Resolutions.} We show the width and height distributions of ground truth animal bounding boxes in the WildLive Benchmark dataset together with a typical animal patch (sampled from the peak). Distributions peak at about 100 pixels and tail off rapidly~(note logarithmic plot scales).}}\vspace{-12pt}
    \label{fig:objectsize}
\end{figure}

\vspace{-3pt}
   \label{sec:dataset}
{\textbf{The WildLive Benchmark Dataset.} Our dataset contains 215,800 bounding boxes and animal segmentation masks along its 291 zebra, giraffe, and elephant tracklets, plus 84 point tracks across 22 UAV-acquired 4K video sequences, totaling 19,139 frames recorded on site at the Ol Pejeta Conservancy in Kenya. Acquisition was conducted via DJI Mavic 3 Enterprise and Pro drones plus a {custom-built quadcopter}\label{refnote} for wildlife missions. Figs.~\ref{fig:dataset} and \ref{fig:objectsize} exemplify frames and key metadata. {Notably, the SAM2 framework is employed to generate segmentation masks by using each verified bounding box as an input prompt, after which all generated masks are synchronised to reconstruct the full frame segmentation}. Overall, the dataset provides verified animal bounding boxes and tracklet IDs, as well as manually corrected, sparse LK pixel trajectories. }

\vspace{-3pt}\section{Method}\vspace{-3pt}
\label{sec:method}
{\textbf{\textbf{Framework Overview.}} As summarised in Fig.~\ref{fig:overview}, our WildLive framework is optimised for near real-time high-resolution video stream tracking onboard drones. It integrates SAHI sampling and YOLO instance segmentation with light-weight sparse optical flow LK tracking and YOLO instance segmentation to localise and follow animals live on UAV platforms. {The framework also incorporates TensorRT optimisation to accelerate inference on Jetson Orin AGX hardware}. The following sections describe the framework and provide technical details of our system.} 

\vspace{-3pt}\subsection{Detection, Segmentation and Point Selection}\vspace{-3pt}
{\textbf{\textbf{Initialisation and Strategic Sampling.}}  High-resolution frames are first processed via SAHI~\cite{sahi} using YOLO11~\cite{yolo11} as detection (and during tracking intialisation also segmentation) model to provide fast localisation of low-resolved animals. This full frame scan~at $t = 0$ is computationally expensive and cannot operate near real-time at high resolutions given today's UAV hardware. Thus, it is used to initiate tracking, whilst later detection and corroboration only focuses on localised 640$\times$640 windows with highest strategic update need. Biological systems focus computational resource in similar ways via foveal vision~\cite{humanvision,humanvisionintracking}. After initiation, two local region categories are prioritised above others for more frequent detection window probing: 
\begin{itemize}
\item \textit{\textbf{Frame Edges:}} to detect new animals entering the frame.
\item \textbf{\textit{Tracked Instances:}} to validate current tracks and to detect disappearances or occlusions.
\end{itemize}}
\noindent {Thus, re-detections are applied to all image windows cyclically, however, at higher temporal sampling rate for above frame regions to focus computational resource.}

{\textbf{\textbf{Sparse Location Selection.}} Selectively processing only point locations on animals rather than performing dense, object-wide, or even frame-wide dense tracking provides order-of-magnitude faster performance to track and thereby interpolate between re-detections. $N$ points \(p_i = [(x,y), id_k]\) at locations \((x,y)\) per animal ID  \(id_k\) are derived as classic Harris corners~\cite{harris} within the animal segment to be tracked where $N=5$ provided reliable performance. We note that with $N>=5$ the system is not sensitive to small changes in $N$. During re-detection validation of tracked animals, these tracked points are resampled to avoid drift, but are required to widely remain within the predicted YOLO segmentation mask to confirm object persistence~(see source code for full details). }

\vspace{-3pt}\subsection {Integration with LK Tracking}\vspace{-3pt}{\textbf{\textbf{Optimised Pyramidal LK Tracker.}} Expanding on early ideas for detector and LK tracker integration~\cite{burghardt2006real} and avoiding frame-by-frame filtered integration for instance via a Kalman Filter~\cite{Kalman,kalmanorg},  we use the pyramidal Lucas-Kanade~\cite{lkp} only as a sparse interpolator between intermittent re-detections to boost speed under a high-resolution regime. Pyramidal processing of the vector of all tracked points~$P^L=[p_i]$ from the coarsest level $L=m$ to the finest resolution level $L=0$ at logarithmic resolution scaling according to \vspace{-8pt}
\begin{equation}
\mathbf{u}^L = \frac{\mathbf{u}}{2^L} , \vspace{-2pt}
\end{equation}
can effectively support rapid high-resolution LK processing of 4K+ imagery. Empirically, a value of $m=5$ yielded fastest overall results without loss of accuracy for our dataset~(see Tab.~\ref{tab:compare4methods}).}

%Specifically, we employ LKP to match selected feature points between two consecutive frames, thereby estimating the motion of the object within the scene. This motion estimation subsequently allows for the prediction of the object's position in the current frame, facilitating continuous and reliable tracking.

%For a given point \( \mathbf{u} \) in frame \( t \), we determine its corresponding location in the subsequent frame by computing the displacement vector \( \mathbf{d} \), which represents the motion of the point between the two frames. This relationship is expressed as  
%where \( \mathbf{v} \) is the estimated position of the point in the next frame.  
%To improve robustness against large displacements in high-resolution images, we used a pyramidal implementation of the Lucas-Kanade method. In this approach, a Gaussian image pyramid is constructed, consisting of multiple levels denoted as \( L = 0, \dots, L_m \), where \( L = 0 \) represents the finest resolution (original image) and \( L_m \) corresponds to the coarsest resolution (most downsampled image).  

{\textbf{\textbf{Linking Point Tracking to Animal Tracklets.}} After determining LK displacement vectors for tracked points, full animal bounding box shifts are predicted as average displacement of the $N$ points associated with an animal instance. Noting that individual LK tracks may fail due to occlusions, aperture limitations, viewpoint changes and more, model drift is possible. To address this, regular re-detections and point re-initialisations on the animal segments are performed, effectively using LK tracking as a data-driven interpolator with minute computational footprint. Performing re-detection updates requires rapid matching of track IDs to re-detected animal masks.}

{\textbf{\textbf{Tracklet Lifecycle and Confidence.}} Let a Point-over-Area (PoA) index be defined as the proportion of tracked points~$p_i$ with~\(id_k\) whose locations~\((x,y)\) fall within a segmentation mask. Matching re-detections and point clouds associated to animal tracklets based on this index together with Intersection over Union (IoU) considerations to address overlap scenarios provides rapid re-association capability between LK point propagations and YOLO re-detections~(see source code for implementation detail). As a result, each YOLO re-detected object is either assigned to an existing animal tracklet or tracked forwards with a new ID if it does not sufficiently match any existing ID. Following~\cite{burghardt2006real}, every tracklet also carries a confidence measure, where in our WildLive system, the measure itself accumulates YOLO re-detection confidence values minus a minimal required confidence per detection over time~(see~\cite{burghardt2006real} for full details). This implements a simple and effective temporal evidence accumulator which, via basic thresholding, controls tracklet termination as well as tracklet validation, that is accepting a tracklet as `1-confident'~(otherwise '0-spurious') and labelling it as such to the user. The latter allows the system to track even 'spurious' YOLO detection candidates with low confidence in order to focus computational resources to validate or dismiss those, whilst only `confident' tracklets are considered for experimentation.}

\begin{table}[t]
\fontsize{9}{10}\selectfont  % Set the font size to 8 points with 9-point line spacing
\centering
\begin{tabular}{c|c|c|c|c}
\toprule
\begin{tabular}[c]{@{}l@{}}\textbf{Tracking}\\ \textbf{Method}\end{tabular}    & \begin{tabular}[c]{@{}l@{}}\textbf{YOLO}\\  \end{tabular} & \begin{tabular}[c]{@{}l@{}}\textbf{fps}\\ \textbf{4K}\end{tabular}   & \textbf{MOTA}            & \textbf{IDF1}             \\
\midrule
\multirow{10}{*}{\begin{tabular}{@{}c@{}}ByteTrack\cite{bytetrack} \\(2022)\\+ SAHI\cite{sahi}\end{tabular}}                                               
 & 8x  & 0.33 & 65.34 & 60.59 \\
 & 8l  & 0.41 & 65.67 & 63.05 \\
 & 8m  & 0.53 & 63.92 & 58.24 \\
 & 8s  & 0.69 & 62.55 & 57.28 \\
 & 8n  & 0.74 & 62.92 & 57.02 \\
 & 11x & 0.31 & 72.50 & 68.44 \\
 & 11l & 0.42 & 67.36 & 61.91 \\
 & 11m & 0.50 & 66.62 & 60.64 \\
 & 11s & 0.62 & 66.52 & 60.70 \\
 & 11n & 0.67 & 62.43 & 55.45 \\
\midrule
\multirow{10}{*}{\begin{tabular}{@{}c@{}}OC-SORT\cite{ocsort} \\(2023)\\+ SAHI\cite{sahi}\end{tabular}}                                                
 & 8x  & 0.33 & 67.98 & 65.96 \\
 & 8l  & 0.43 & 67.69 & 66.35 \\
 & 8m  & 0.55 & 62.11 & 59.35 \\
 & 8s  & 0.74 & 53.38 & 54.35 \\
 & 8n  & 0.79 & 47.62 & 48.15 \\
 & 11x & 0.31 & 70.82 & 67.90 \\
 & 11l & 0.44 & 64.19 & 63.63 \\
 & 11m & 0.52 & 62.94 & 61.00 \\
 & 11s & 0.66 & 56.90 & 55.75 \\
 & 11n & 0.71 & 50.29 & 49.90 \\
 \midrule
\multirow{10}{*}{\begin{tabular}{@{}c@{}}SORT\cite{sort} \\(2016)\\+ SAHI\cite{sahi}\end{tabular}}                                                      
 & 8x  & 0.33 & 74.41 & 55.69 \\
 & 8l  & 0.43 & 74.77 & 57.55 \\
 & 8m  & 0.56 & 68.28 & 51.23 \\
 & 8s  & 0.74 & 61.12 & 48.89 \\
 & 8n  & 0.79 & 57.93 & 44.14 \\
 & 11x & 0.31 & 75.83 & 60.12 \\
 & 11l & 0.46 & 71.08 & 55.60 \\
 & 11m & 0.51 & 68.47 & 56.10 \\
 & 11s & 0.66 & 66.31 & 50.31 \\
 & 11n & 0.71 & 59.60 & 45.91 \\
\midrule
\multirow{10}{*}{\begin{tabular}{@{}c@{}}\textbf{WildLive}\\ \textbf{(Ours})\end{tabular}}  
 & 8x  & 4.79          & 76.65          & 75.86          \\
 & 8l  & 5.29          & 78.70          & \textbf{79.03} \\
 & 8m  & 5.60          & 75.51          & 74.18          \\
 & 8s  & \underline{ 5.78}    & 77.15          & 76.46          \\
 & 8n  & 5.68          & 70.29          & 69.31          \\
 & 11x & 5.72          & \textbf{81.17} & \underline{79.02}    \\
 & 11l & 5.12          & \underline{81.02}    & 78.23          \\
 & 11m & 5.30          & 77.74          & 78.71          \\
 & 11s & 5.76          & 75.52          & 74.76          \\
 & 11n & \textbf{6.32} & 75.07          & 73.45          \\   
\bottomrule
\end{tabular}\vspace{-5pt}
\caption{\tb{\textbf{Comparative WildLive System Benchmarks (on Tesla GPU).} MOT evaluation of processing speed in frames per second (fps) and MOTA/IDF1 measures ($\%$) for one full run across the WildLive Benchmark Dataset benchmarked at full 4k resolution on a Tesla P100-PCIE-16GB GPU. Note order-of-magnitude speed advantages achieved by utilising LK tracking as temporal interpolator between re-detections. Different YOLO versions are probed, with the YOLO11n, for instance, having a 2.9M parameter resource footprint~\cite{ultralytics}.}}\vspace{-13pt}
\label{tab:compare4methods}
\end{table}

\vspace{-5pt}\section{Experiments}\vspace{-5pt}

\begin{comment}
    
\tb{\textbf{Experimental Setup.} We evaluate WildLive on offline Tesla P100-PCIE-16GB GPU hardware~(see Fig.~\ref{tab:compare4methods}) \dat{(see Tab.~\ref{tab:compare4methods})} and pinpoint its speed on the Jetson Orin AGX onboard GPU environment~(see Fig.~\ref{tab:tradeoff}) \dat{(see Tab. \ref{tab:tradeoff} and Tab. \ref{tab:hdspeed}}). We utilise standard networks (e.g. YOLO, SAHI as cited) for Tesla experiments, but based on these perform TensorRT optimisation of networks for maximum efficiency during Jetson experiments sharing\footref{note2} all optimised network weights.  Jetson deployment is further facilitated via Docker containerisation of the system offering deployment flexibility and scalability for the community. Successful test flights in Kenya~(shown in Fig.~\ref{fig:overview}) constitute a physical proof-of-concept that the full WildLive system can indeed operate on a custom built, flight-tested~UAV\footref{note2}.}

\end{comment}

{\textbf{Experimental Setup.} We evaluate WildLive offline on Tesla P100-PCIE-16GB GPU hardware~(see Tab.~\ref{tab:compare4methods}) and pinpoint its speed on the Jetson Orin AGX onboard GPU environment~(see Tab. \ref{tab:hdspeed} and Tab. \ref{tab:tradeoff}). For the Tesla experiments, we utilise standard YOLO networks combined with SAHI~\cite{sahi} for detection and segmentation, providing baseline performance. In contrast, for Jetson deployment, we perform extensive TensorRT optimisation on these same networks to maximise inference speed and efficiency, publishing optimised network weights\footref{note2}. Jetson deployment is further facilitated via Docker containerisation, offering flexibility and scalability for community use. All experiments are conducted with the Jetson Orin AGX running in maximum performance mode. Successful test flights in Kenya~(depicted in Fig.~\ref{fig:overview}) constitute a physical proof-of-concept that the full WildLive system can indeed operate on a custom built, flight-tested~UAV\footref{note2}.}

\begin{comment}
\begin{table*}[h]

\centering

\begin{tabular}{|c|c|c|c|c|c|c|c|}
\hline
Dataset & Videos & Frame rate & Resolution &  Real world & Long video  & Animal domain \\
\hline
TAP-Vid DAVIS first \cite{tapvid} & 30 & 24 & 1080 x 1920 & Yes & No  & Yes \\
PointOdyssey \cite{pointodyssey} & 159  & 30 & 540 x 960 & No & Yes  & Yes\\
RoboTAP \cite{robotap} & 265 & - & 256 x 256 & Yes & No  & No \\
Kubric \cite{kubric} & -  & 8 & 256 x 256 & No & No  & No \\
BADJA \cite{badja} & 9  & 24 & 1080 x 1920 & Yes & No  & Yes \\
\textbf{Ours} & 22 & \textbf{30} & \textbf{3840 x 2160} & \textbf{Yes} & \textbf{Yes} & \textbf{Yes}\\

\hline
\end{tabular}
\caption{Comparison of point tracking datasets our dataset is the highest resolution, frame rate, real-world data, has long videos and includes trajectories of points on animal body. }
\label{tab:datacompare}
\end{table*}
\end{comment}
\section{Results}\vspace{-5pt}

\tb{\textbf{Comparative System Benchmarks.} We compare WildLive against recent SOTA~\textit{full object trackers} computationally suitable for edge device deployment, in particular ByteTrack~\cite{bytetrack}, OCSORT~\cite{ocsort}, and SORT~\cite{sort}. {To support these trackers with detection input, we use the SAHI~\cite{sahi} technique rather than relying solely on a standard YOLO model.} We benchmark processing speeds together with both Multi Object Tracking Accuracy~(MOTA) and Identification F1~(IDF1) scores. The latter complements MOTA regarding limitations on how long trackers correctly identify objects. Full results are shown in Tab.~\ref{tab:compare4methods} and confirm order-of-magnitude gains in processing speed compared to other tested techniques resting on only intermittent re-detection bridged by computationally negligible LK point tracking. {As shown in Tab.~\ref{tab:compare4methods} and Tab.~\ref{tab:hdspeed}, WildLive achieves a processing speed approximately 10 times faster than that of the other trackers mentioned above and attains near real-time performance (up to 17.81 fps for HD).} Accuracy, maybe surprisingly, slightly improves too, leading to best MOTA at $81.17\%$ and IDF1 at $79.03\%$: utilising PoA \textit{and} IoU together for ambiguity resolution proves superior compared to the IoU-centred full object tracking methods tested which have no direct access to segmentation masks.}
\begin{comment}
    
\tb{\textbf{Runtine Speed Estimation.} In order to estimate performance on drone hardware specifically, we also benchmark WildLive's TensorRT-optimised processing speed on Jetson hardware as shown in Tab.~\ref{tab:tradeoff} for a range of re-detection regimes scaling from our standard system~(single window re-detection per frame named \textbf{Ours}) at 7.53fps down to the base case of permanent full frame re-detection at 2.45fps. We finally note that \textbf{Ours} processes HD streams at 17fps+. }
\end{comment}

{\textbf{Runtime Speed Estimation.} The WildLive system employs TensorRT optimisation on Jetson hardware to boost real-time performance, replacing the default PyTorch model during deployment. We evaluate processing speed on drone hardware through two sets of benchmarks. First, we assess the inference speed of all supported detection models on both HD and 4K video streams; on HD input, for instance, WildLive achieves up to 17.81 fps with a YOLO11n detector (see Tab.~\ref{tab:hdspeed}). Secondly, we measure inference speed across a range of re-detection regimes, from our standard system (single-window re-detection per frame) at 7.53 fps down to the base case of permanent full-frame re-detection at 2.45 fps, using the fastest detection model, YOLO11n (as reported in Tab.~\ref{tab:tradeoff}). These benchmarks show that TensorRT optimisation yields further improvements in inference speed, although it roughly doubles the model size.}

\begin{table}[h]
\fontsize{9}{10}\selectfont  % Set the font size to 8 points with 9-point line spacing
\centering
\begin{tabular}{c|c|c|c|c}
\toprule
Yolo & \begin{tabular}[c]{@{}l@{}}fps on 4K \\ (TRT)\end{tabular} & \begin{tabular}[c]{@{}l@{}}fps on HD\\ (TRT) \end{tabular} & \begin{tabular}[c]{@{}l@{}}fps on 4K \\ (Pytorch)\end{tabular} & \begin{tabular}[c]{@{}l@{}}fps on HD\\ (Pytorch) \end{tabular} \\
\midrule
8x   & 6.09          & 11.75          & 5.62          & 8.56           \\
8l   & 6.97          & 14.27          & 6.02          & 11.69          \\
8m   & 7.03          & 16.44          & 6.42          & 13.55          \\
8s   & 7.21          & \underline{17.76}    & 6.80          & 14.66          \\
8n   & 7.17          & 17.69          & \underline{6.88}    & 14.77          \\
\midrule
11x  & 5.94          & 11.02          & 5.75          & 8.51           \\
11l  & 6.21          & 15.63          & 5.89          & 11.32          \\
11m  & 7.08          & 16.89          & 6.31          & 13.33          \\
11s  & \underline{7.30}    & 17.57          & 6.49          & \underline{15.92}    \\
11n  & \textbf{7.53} & \textbf{17.81} & \textbf{6.96} & \textbf{16.79} \\
\bottomrule                     
\end{tabular}\vspace{-6pt}

\caption{{\textbf{WildLive Speed Across YOLO Versions and Video Resolutions.}  
Speeds are reported for both 4K and HD input resolutions on Jetson AGX Orin. The system achieves its highest performance with YOLO11n at 17.81 fps on HD data using our single-window re-detection approach and TensorRT optimisation.}}

\label{tab:hdspeed}
\end{table}

\begin{table}[h]
\fontsize{9}{10}\selectfont  % Set the font size to 8 points with 9-point line spacing
\centering
\begin{tabular}{l|c}
\toprule
\begin{tabular}[c]{@{}l@{}}\textbf{Re-Detection Windows}\\ (per 4K frame time step)\end{tabular} & \begin{tabular}[c]{@{}l@{}}\textbf{fps}\\  (Jetson)\end{tabular}             \\
\midrule
01 (Standard System (Ours))                                                               & \textbf{7.53}                                                                 \\
02                                                               & \underline{6.94}                                                                                    \\
04                                                               & 5.87                                                                                       \\
08                                                               & 4.59                                                                                   \\
16                                                              & 3.28                                                                                       \\
24 (Full Frame Re-detection)                                                             & 2.45                                                                                                     \\
\bottomrule
\end{tabular}\vspace{-6pt}
\caption{{\textbf{WildLive Speed Benchmarks for Different Re-Detection Window Configurations.} System speed in frames per second (fps) on the WildLive Benchmark Dataset, measured across different numbers of re-detection windows per 4K frame. Inference was performed using a TensorRT engine configured based on the number of re-detection windows. WildLive with permanent full frame re-detection performs at $2.45$ fps, whilst single window probing~(Standard System (Ours)) allows for $7.53$~fps.}  }
\begin{comment}
\caption{\tb{\textbf{WildLive Speed Benchmarks (Jetson).}  System speed for one full run across the WildLive Benchmark Dataset given in frames per second (fps) for our TensorRT optimised Jetson AGX Orin onboard platform across different numbers of re-detection windows probed per 4K frame. Note that, WildLive with permanent full frame re-detection performs at $2.45$fps, whilst single window probing~(Ours) allows for $7.53$fps. Benchmarks include an avg. $12\%$ improvement achieved by TensorRT optimisation.} }
\end{comment}

\label{tab:tradeoff}
\end{table}

\vspace{-3pt}\section{Conclusion and Future Work}\vspace{-3pt}
We conclude that WildLive's high-resolution, high-speed tracking capabilities are close to those needed to fuel seamless, animal-reactive drone navigation based on video processing directly onboard UAVs, albeit accuracies under a small object regime still leave significant room for detection and tracking improvements. Such availability could revolutionise the way UAV conservation missions are conducted in terms of range, cost, ease of use, and mission type.

{{\textbf{\textbf{Improving Evaluation, Small Object Detection and Segmentation.}}}
To address limitations, we plan to develop a fast and efficient instance segmentation model specifically tailored for detecting and segmenting small animals, adapting it to the unique challenges of UAV-based wildlife monitoring. Additionally, we are in the process of benchmarking the system with TETA~\cite{teta}, testing alternative detectors including RT-DETR~\cite{rt-detr}, and producing full speed-accuracy benchmarks that are statistically more robust using multiple runs and sampling widely from the WildLive parameter space, with rigorous field testing of a prototype.}

{{\textbf{\textbf{Long Term Goal.}}} 
Ultimately, our goal is to integrate these computer vision capabilities with UAV navigation, enabling fully autonomous, vision-driven BVLOS wildlife monitoring missions. This may allow to address currently missing conservation capabilities including autonomous monitoring of large wildlife reserves.}

\begin{comment}
\tb{We conclude that WildLive's high-resolution, high-speed tracking capabilities are close to those needed to fuel seamless, animal-reactive drone navigation based on video processing directly onboard UAVs, albeit accuracies under a small object regime still leave significant room for detection and tracking improvements. Such availability could revolutionise the way UAV conservation missions are conducted -- in terms of range, cost, ease of use as well as mission type.\dat{To support this, we will develop a fast and efficient model specifically designed to detect and segment small objects, adapting it to the unique demands of UAV-based wildlife monitoring. Besides,} we are in the process of producing full speed-accuracy benchmarks for multiple runs across the parameter space of WildLive and envisage rigorous field testing of the prototype next.\dat{ To further enhance inference efficiency, we plan to optimize the TensorRT model by quantizing it to int8 instead of fp32, which is expected to significantly reduce computational load and improve runtime performance on embedded hardware. This optimization will reduce computational demands and enable faster, more efficient deployment on drone platforms.} Eventually, we will link our computer vision methods directly to UAV navigation for vision-driven autonomous BVLOS wildlife monitoring missions.}

\end{comment}

\section*{Acknowledgements}\vspace{-5pt}
{This work was supported by the WildDrone project (under the
Marie Skłodowska-Curie Actions (MSCA) - grant agreement ID: 101071224) and UK Research and Innovation (EPSRC/UKRI - project reference: EP/X029077/1), the Imageomics Institute (NSF HDR Award 2118240), and ICICLE (NSF grant OAC-2112606).}

\section*{Ethics Statement}\vspace{-5pt}
\label{sec:ethics}
When collecting our WildLive Benchmark Dataset and during test flights, we adhered to strict ethical standards to ensure the wellbeing of the animals and the appropriateness and integrity of the research. Our data was collected using drones in the Ol Pejeta Conservancy, Kenya. 
%Great care was taken to minimise any potential disruption to wildlife in general and filmed animals in specific. 
Drones were operated at safe distances to avoid causing stress or disturbance to animals. Additionally, all recordings were made in a non-invasive manner, with no direct interaction with wildlife. To further protect privacy and adhere to ethical guidelines, we ensured that human faces are not contained in  recordings, focusing solely on animals and their behavior. All data collection procedures followed relevant local regulations, such as those provided by the Kenya Civil Aviation Authority (KCAA)(Authorization number KCAA/UAS/OPS/0068/2025), the Kenya Wildlife Research and Training Institute (KWRTI), and other ethical guidelines. Efforts were made to respect the natural behaviours of the animals maximally. By prioritising animal welfare, privacy, and ethical research practices, the dataset can contribute to scientific advancements while minimising any harm to the environment and its inhabitants.

%%%%%%%%% REFERENCES

{
    \small
    \bibliographystyle{ieeenat_fullname}
    \bibliography{main}
}
\end{document}